# Comparative Study of MPPT and Parameter Estimation of PV Cells using ANN

*

**Sahil Kumar**

Vellore Institute of Technology
Jamshedpur
Sahil.kumar2019a@vitstudent.ac.in

**Vajayant Pratik**

Vellore Institute of Technology
Patna
Vajayant.Pratik2019@vitstudent.ac.in

**Sahitya Gupta**

Vellore Institute of Technology
Jhansi
Sahitya.gupta2019@vitstudent.ac.in

**Pascal Brunet - Gabriel Dabadie**

NALDEO Technologies Industries
France
Pascal.brunet@naldeo.com

## Abstract

The presented work focuses on utilising machine learning techniques to accurately estimate accurate values for known and unknown parameters of the PVLIB model for solar cells and photovoltaic modules. Finding accurate model parameters of circuits for photovoltaic (PV) cells is important for a variety of tasks.

An Artificial Neural Network (ANN) algorithm was employed, which outperformed other meta-heuristic and machine learning algorithms in terms of computational efficiency. To validate the consistency of the data and output, the results were compared against other machine learning algorithms based on irradiance and temperature. A Bland Altman test was conducted that resulted in more than 95 percent accuracy rate. Upon validation, the ANN algorithm was utilised to estimate the parameters and their respective values.

## 1 Introduction

Solar energy has been developed as a better alternative to fossil fuels in the past few years. It is a renewable and infinite source of energy which does not have a bad impact on the environment. It is also cheap and easily accessible, making it a better alternative for both personal and commercial purposes.

Solar Arrays are made when PV modules used in solar panels are connected together. Energy is produced when sunlight falls on Solar Panels which can be used instead of Fossil fuel's produced energy.

For execution of a PV system under different situations, estimating the parameters in a PV model plays an important role because it enables us to optimise the design and performance of the system which leads to increased energy production and improved performance.

If a PV system is not performing as expected, then identification of parameters of the PV model helps identify the root cause of the problem. This could be due to factors such as shading, module mismatch, or degradation over time. By accurately estimating the parameters, we can determine the best method to improve its performance.

---

*<u>Citation</u>: **Comparative Study of MPPT and Parameter Estimation of PV Cells using ANN**

In addition, accurately estimating the Parameters in a PV model is important for financial modelling and energy yield predictions. By knowing the expected energy production of a PV system, investors and project developers can make informed decisions about the feasibility and profitability of a project.

## 2 PV Model and Maximum Power Point

### 2.1 PVLib Library

The single diode model is a widely applied model for simulating PV cell's electrical behaviour under different environmental conditions, such as temperature and irradiance. The single diode equivalent diagram of PV cells is given in figure below[1].

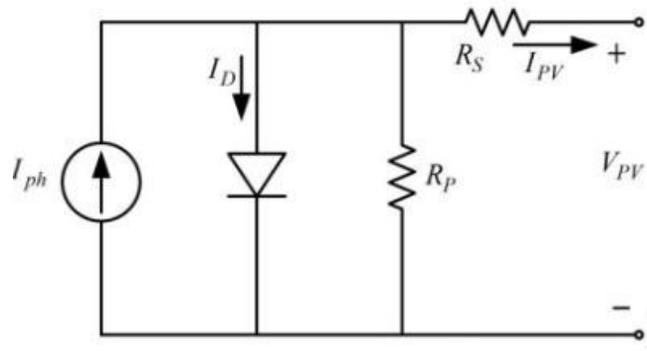

Figure 1: Single diode circuit model

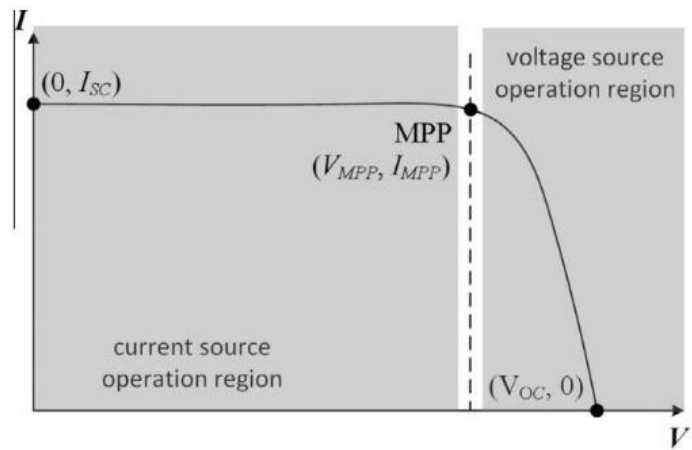

Figure 2: Current, Voltage and Power points in I-V curve

PVLIB's implementation of the single diode model includes several parameters such as the diode ideality factor, series resistance, and shunt resistance. With the PVLIB library, one can quickly and precisely simulate the behaviour of PV modules under different environmental conditions and examine the performance of their PV systems.

Some of the tasks that can be performed with PVLib include:

- Calculating the solar position and solar radiation on a given location and date/time

- Modelling the performance of PV modules under different conditions (temperature, irradiance, shading, etc.)

- Estimating the performance of a full PV system based on its design and location

- Simulating the output of a PV system over time, taking into account weather data and other factors

## 2.2 PVLib based PV model

With the help of PVLib library, a PV system Model is built to simulate it in any location by giving the input of timezone, altitude, latitude and longitude.

For calculating the peak power for an instant, the value of solar position, irradiance, temperature and wind speed is needed.

Solar position is determined using PVLib function

> *Solar Angle = pvlib.solarposition.getsolarposition(time, latitude,longitude, altitude)*

Using the data of solar Angle , total irradiance is calculated which is a prerequisite for formulating the equation for Maximum power point. Total irradiance is a combination of direct and diffuse irradiance. The calculation uses soiling and `iam_b` as Modal parameters that impacts the output of the model.

> Pmpp = pvlib.pvsystem.singlediode(pvlib.pvsystem.calcparams-desoto( Totalirradiance .pvlib.temperature.pvsyst-cell(Total irradiance , temp,wind ,*u-c ,u-v, eta-m , alpha-absorption)*, alpha-sc, a-ref, I-L-ref , I-o-ref , R-sh-ref , R-s,EgRef , dEgdT)).p-mp/1000]no-of-module* (1-*mismatch)(1-wiring)(1-connection)(1-lid)(1-nameplate-rating)*

### 2.2.1 Nomenclature

These are the list of parameters that are used to model the electrical and optical characteristics of a PV module:

| Parameter name | Description |
|---|---|
| alpha-sc | short-circuit current temperature coefficient |
| a_ref | measure of the degree of recombination in the PV cell |
| I_L_ref | light-generated current |
| I_o_ref | reverse saturation current |
| R_sh_ref | shunt resistance |
| R_s | Series resistance |
| EgRef | Band Gap energy of the semiconductor |
| dEgdT | Temperature coefficient of the bandgap energy |

Following are exhaustive list of parameters that impacts the accuracy of model:

| Parameter name | Description |
|---|---|
| lid | Light Induced Degradation on the degradation of modules in the first hours of exposure |
| u_c | Combined heat loss coefficient |
| u_v | Combined heat loss factor influenced by the wind |
| eta_m | External output of the module |
| iam_b | Adjusts the incidence angle modifier as a function of angle of incidence |
| albedo | Albedo of the location |
| wiring | Ohmic loss (percent) of cables between panels |
| soiling | Coefficient that characterises fouling losses |
| mismatch | differences in performance between individual photovoltaic cells |
| connection | Ohmic loss (in percent) of the contact resistance between the panels |
| alpha-absoption | Absorption coefficient |

Adjusting the values of above parameters, changes the output impactfully. Understanding the parameter's intrinsic property and its role in the PV module is essential.

**Albedo:** Defines the reflective properties of a surface. The ratio of albedo is the amount of incoming radiation that is reflected by the surface to the total amount of incoming radiation.
The value of Albedo ranges between (0,1), in which a higher value means a greater degree of reflectivity.

**Light-induced degradation (LID):** When a new solar cell is exposed to sunlight, LID causes a temporary reduction in its production. The formation of defects in the crystal structure of the cell causes reduction.

**Soiling:** The performance of PV cells reduces due to formation of dirt, dust and other particles on the surface of PV modules. This is referred to as Soiling.

**Wiring:** This parameter refers to the electrical characteristics of the cables, connectors, and other components used to connect PV modules to inverters, batteries, and other system components

**Alpha absorption:** It calculates the spectral response of PV cells. It is a measure of the degree to which a material absorbs light of a given wavelength or energy level. The alpha absorption value is defined in units of $cm^{-1}$ or $m^{-1}$.

**Mismatch:** It is a difference in performance between individual PV cells in a PV system. Mismatch occurs due to several factors, including manufacturing variations, shading or soiling, and differences in the operating conditions of individual cells or modules.

This PVLib model of Solar cell is used to obtain the Maximum Powerpoint of the system at an instant. MPP tracking is essential for the efficiency of PV Module.

In this paper, ML algorithms such as Random Forest and SVR(Support Vector Regression) are applied to understand the PV model in a better way. It is also essential to get insight on data( Irradiance, temperature, etc) that helps in setting appropriate hyperparameters in parameter estimation of the PV model.

## 3 Maximum Power Point Tracking

### 3.1 Overview on MPPT

The previous work on MPPT has led to the evolution of many algorithms that have significantly refined the accuracy and efficiency of PV systems. Various techniques such as Perturb and Observe, and Hill Climbing, Incremental Conductance, for MPPT[2] in PV systems. In recent years, researchers have also focused on applying ML algorithms, such as Random Forest and SVR, for MPPT.

The comparative analysis of these algorithms can provide valuable insights into their performance and suitability for different applications. This study builds on the previous work on MPPT and presents a comparative analysis of Random Forest and SVR algorithms for MPPT in a real-world PV system. The findings of this study can provide a valuable reference for future research on MPPT in PV systems.

The two primary parameters that determine the operating point of a PV panel are solar irradiation and temperature. These are the reasons for determining the maximum power point (MPP) of the system.

Solar irradiation is the quantity of sunlight that reaches the solar array, and it directly affects the amount of electrical power built by the system. The more will be the solar irradiation, the more electrical energy the PV panel produces. Temperature also plays an important job in finding the MPP of the PV system.

In PV modules,when the temperature increases, the voltage decreases, while the current increases. This phenomenon is known as the temperature coefficient of the PV module. Thus, at a higher temperature, the MPP shifts to a lower voltage and higher current. Conversely, at lower temperatures, the MPP shifts to a higher voltage and lower current.

Overall, the V-I characteristics of a Solar Cell are primarily influenced by solar irradiation(G) and temperature(T) .

### 3.2 Random Forest

The results obtained from the Random Forest algorithm are better than decision tree classification and other linear methods. Random Forest is a structured algorithm for predicting output in classification problems. Moreover, compared to other ensemble techniques, Random Forest produces more accurate and stable results.

The Random Forest algorithm has been successfully applied to various real-world problems such as improving rainfall rate and resampling field spectra. Previous studies have demonstrated that the Random Forest model has higher robustness, stability, and success rates than other models if the model parameters are appropriately tuned.

Furthermore, previous approaches for modelling Random Forest for regression problems have been successful as well. Therefore, executing the Random Forest model for creating a MPPT model for a PV system can lead to an accurate and better outcome compared to other models.

In this paper, we have designed and implemented a Random Forest model to track the maximum power point of a PV module accurately and efficiently at an instant. The prototype is modelled using Python and its libraries.

#### 3.2.1 Proposed Random Forest MPPT Approach

The method proposed for MPPT uses irradiance and temperature as inputs, unlike other conventional approaches. This approach is suitable since irradiance and temperature are commonly used factors to determine the PV module related result.

Key factors of RF for regression:

- Random forest is an ensemble learning method that uses bootstrap sampling to create multiple decision trees. Each bootstrap sample contains approximately one-third of the original dataset, with the remaining data forming the out-of-bag (OOB) samples. OOB samples are used to validate each decision tree's accuracy without cross-validation. By combining multiple decision trees, random forest produces more accurate predictions and reduces the risk of overfitting.

- Random forest algorithm builds a regression tree for each bootstrap sample drawn from the original dataset. Illustrated in Figure 3.

At each node, the predictor variable with the least RMSE is selected for the division. This selection of variable is used to identify the best split for each node.

The OOB error is determined by comparing the above explained predicted values with the reference values using the data that were not included in the bootstrap samples. These OOB elements can be used to evaluate the performance of each regression tree and provide a rough average of the effectiveness of the RF model.

To advance the effectiveness of the RF model, two important parameters need to be adjusted: ntree and total data samples. Using OOB data, the tuning process is conducted to find the best values for ntree and m that result in the best model accuracy while avoiding overfitting.

#### 3.2.2 Overview of Random Forest Training

The data on irradiance and temperature is used to train a model. To achieve this, experimentation is done with different values of the algorithm's hyperparameters to optimise its performance.

Comparative Study of MPPT and Parameter Estimation of PV Cells using ANN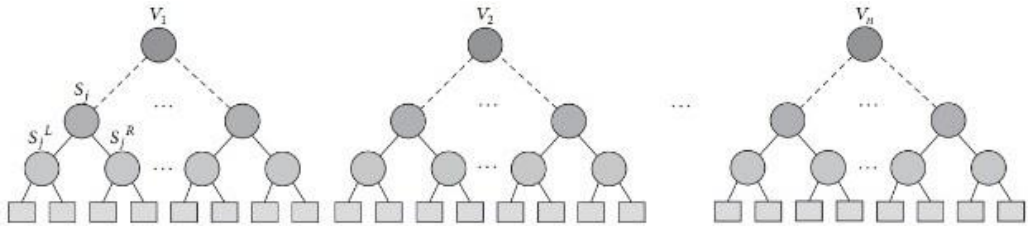

Figure 3: Random Forest

One of the hyperparameters considered was the "max_features" parameter, which determines the number of features to consider when dividing a node in the Random Forest algorithm. Three different values for this parameter: "sqrt", "log2", and the default value (which considers all features) is considered. Trained and tested the model using each of these values and compared their performance.

Another hyperparameter that is experimented with was the number of trees in the forest. The out-of-bag (OOB) score was plotted, figure 4, which is a measure of the model's accuracy, against the number of trees to determine the minimum OOB score and corresponding number of trees required to reduce computation time.

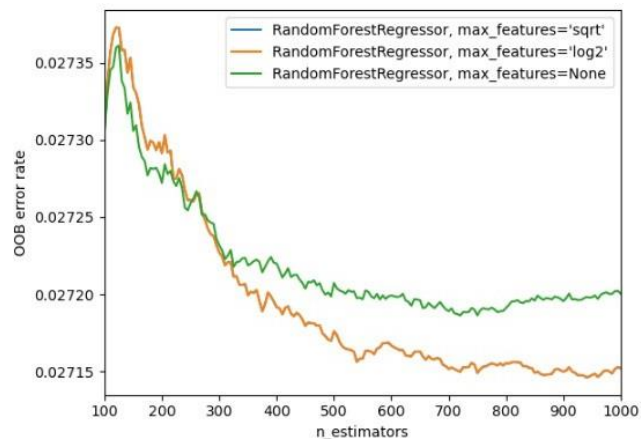

Figure 4: OOB error rate v/s no. of tress

After performing training with various 'max_features' values and analysing the OOB score plot, it was discovered that the lowest OOB error rate occurred when max_features was set to 'log2'. It was also observed that the error rate decreased significantly after 500 trees and remained stable afterward.

Based on these findings, we selected a number of trees equal to 700 and max_features equal to 'log2' to train the model. This combination of hyperparameters resulted in the lowest error rate compared to other max_features values. Training the model using the above parameter gave significant output with high accuracy in predicting the MPP of testing data.

### 3.3 Support Vector Regression

A machine learning approach called Support Vector Regression (SVR) is utilised for regression analysis. It is founded on the widely known Support Vector Machine (SVM) method, which is utilised for classification jobs.
For regression issues, it employs the same basic idea as SVM.
Support vector machines (SVM) search in multidimensional space for a line or hyperplane that divides these two classes.

In order to forecast a continuous target variable properly given a collection of input features, SVR seeks out the best fit line. The algorithm finds a group of support vectors that serve as the dividing line between various classes, and then it uses these support vectors to determine[6].

### 3.3.1 Overview of SVR

The target while traversing, is to select the data points from the set that are inside the boundary line's decision.
The major benefit of SVR is that it can cope with non-linear connections between the input and the target variables, by using kernel functions. This allows SVR to model complex relationships that may not be possible with linear regression models.

There are 4 parameter in SVR:

- Kernel: A linear separation hyperplane is simpler to locate in a higher-dimensional feature space created by the use of a kernel, which is employed to transfer the initial input space into that space. There are numerous varieties of kernels, such as radial basis, polynomial, and linear kernels. Because it can have a considerable impact on the SVR's performance, it is crucial to choose the right kernel function carefully.

- Maximum error eps: It establishes the tolerance for mistakes in the prediction. The breadth of the insensitive zone, also known as the tube, the region around the hyperplane where errors are permitted, is determined by the value of epsilon. Any data point within this zone is considered to be correctly predicted, and the algorithm does not try to optimise its position.

- The regularisation parameter C: It controls the trade-off between the margin and the error of the model and also determines the amount of penalty that the model incurs for violating the margin, i.e., for misclassifying a data point.

- The kernel coefficient parameter gamma: It controls the shape and smoothness of the kernel function.

### 3.3.2 Proposed Support vector regression MPPT approach

In the beginning, linear regression is applied to the provided dataset after cleaning it and a graph is plotted showing the prediction. But linear regression is not applicable because of non-linear relationships between parameters, hence SVR can be applied.
Hence, to find the best values of the parameters of SVR, the kernel is first set as linear, and the best values of C and epsilon are calculated. The value of C is then varied within a range to identify the value that yields the highest percentage of data points within epsilon outside the regression line.
To find the best values of C and epsilon, two scoring criteria, MAE and percentage within epsilon, are considered using the grid search method. At each point, the MSE is determined, which reduces as the percentage within epsilon increases.
After applying these methods the best value of C obtained is 9.38 and best value of eps is 2.52. Illustrated in figure 6. Hence percent of data within epsilon obtained is 93.07 which is maximum with respect to all the above methods.

Table 3. The value of C and eps and the corresponding percent within epsilon obtained from different methods

| C    | Eps  | PercentWithinEps | MAE   |
|------|------|------------------|-------|
| 0.01 | 5    | 78.97            | 4.570 |
| 1    | 5    | 92.30            | 4.223 |
| 8.16 | 5    | 92.42            | 4.229 |
| 6.53 | 0.01 | 92.64            | 2.540 |
| 9.53 | 2.25 | 93.07            | 3.547 |

After obtaining the best values of C and epsilon, the MSE is calculated for different types of kernel functions. It was observed that the radial basis function (RBF) kernel yielded the lowest MSE. The best gamma value can also be determined using the grid search method.

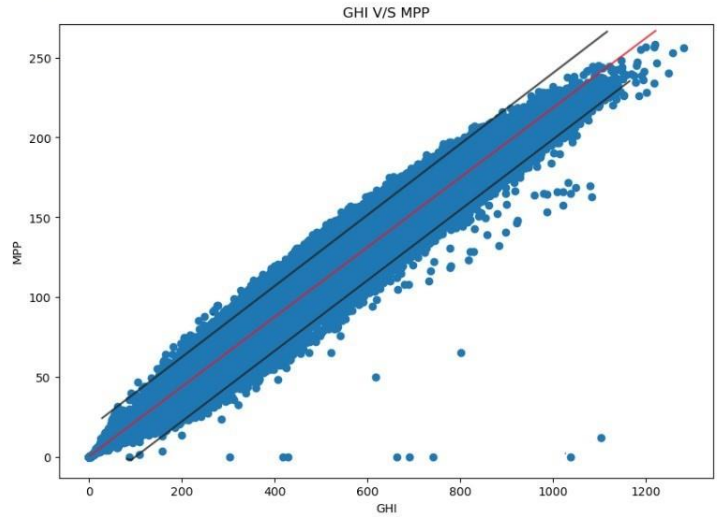

Figure 5: Scatter plot of GHI vs MPP with Epsilon as 5

In the Figure 5, 78 percent of the data points are within the epsilon range, it comprehends as 78 percent of data is fairly tested and trained.

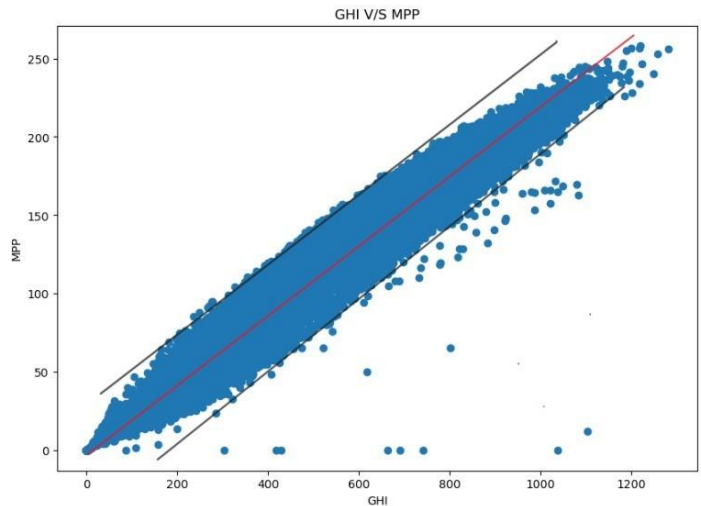

Figure 6: Scatter Plot of GHI vs MPP using Grid Search Method

Whereas, in Figure 6, after using the Grid search method, 93 percent of the data points are within the epsilon range, it comprehends as 93 percent of data is fairly tested and trained.

# 4 Parameter Estimation and Tuning

Parameter estimation and tuning is needed to optimise the working of PV modules. Determining and tuning the important parameters precisely that affects the performance of the solar module is a major task while designing the system. This is to maximise energy production, improve efficiency, and enhance its reliability and durability[9].
This ensures that the module operates at its maximum power point (MPP) and within its design limits, reducing energy losses and the risk of damage or failure.

## 4.1 Need of Parameter Estimation

While MPPT and parameter estimation are both used in the design and control of PV systems, MPPT is primarily used for optimising the power output of the system, whereas parameter estimation plays a role in designing the behaviour of the solar module and designing control strategy for regulating the output voltage or current.
By estimating these parameters, we can create mathematical models that describe the cell's behaviour under different conditions. So to optimise the power output and efficiency of the system, these mathematical models can be used for design and control of PV such as inverter control and MPPT algorithm systems. Parameter estimation is also useful for diagnosing faults or defects in PV cells, which can affect their performance and lifespan.

## 4.2 Methods applied for Parameter Estimation

Methods such as the curve-fitting method, the incremental conductance method, and the perturb and observe method are used for parameter estimation and tuning of PV cells. Even though these methods have been applied for several real world problems, they are not as accurate as ML methods for several reasons.

**The Curve-fitting method** is used for identifying and calculating the parameters of a PV module. The algorithm requires prior knowledge of the PV cell characteristics. The model is then iteratively adjusted until it matches the measured I-V curve.

However, this method may not fully capture the complexity of PV cells due to its limitations. The mathematical models of this method may not be able to capture nonlinear and dynamic behaviour accurately. For determining the accuracy of the model, quality of initial assumption of model parameters is important, which can be difficult to obtain. Therefore, the curve-fitting method may not be as accurate or robust as ML methods.

**The Incremental conductance method** is a more advanced technique for parameter estimation. The method involves measuring the V and I of the Solar cell and determining the incremental conductance, which is then compared to zero to determine the MPP.

However, the incremental conductance method may not be as accurate as ML methods. The method relies on an assumption that the PV cell operates at the MPP when the incremental conductance is 0, which may not be true.
In addition, the method is easily impacted by environmental circumstances, such as irradiance and temperature, which can impact the accuracy of the estimated parameters.
Overall, traditional algorithms for parameter identification of PV modules have limitations and may not be as accurate or robust.
Soft computing methods work better than traditional methods of parameter estimation. They are a family of computational techniques that are used for various applications, including parameter estimation of PV cells.

**Fuzzy logic** is a mathematical framework that deals with uncertainty and imprecision. Fuzzy sets are sets that can have partial membership[10]. Fuzzy logic can be used for parameter estimation of PV cells by defining fuzzy rules based on the input-output relationships of the PV cell. The performance of fuzzy logic in parameter estimation depends on the accuracy of the defined rules and the quality of the available data.

**Genetic algorithms** are algorithms that are driven by the method of natural selection. It can be applied for parameter extraction of PV modules by stating a fitness function that evaluates the functionality for different parameter values. Generic algorithm then looks for the best set of parameters that maximises the fitness function. The quality of the fitness function and the available data is a key factor for the performance of genetic algorithms in parameter estimation.

**Particle swarm optimization** can be applied for parameter identification of PV modules by stating a fitness function and a population of particles that represent different parameter values. The algorithm then searches for the optimal set of parameters by updating the position and velocity of each particle based on its previous journey and the knowledge of the optimal particle in the population[11]. The performance of particle swarm optimization in parameter estimation depends on the quality of the fitness function, the population size, and the available data.

When comparing the performance of soft computing methods with ML methods, such as ANNs, it is important to consider the specific problem and the available data.

In ML methods, ANN have gained popularity in recent years due to their ability to learn from data and handle non-linear relationships [8]. ANNs have shown promising results in parameter estimation of PV cells, especially when dealing with large and complex datasets and learning from data without the need for prior knowledge or assumptions about the PV cell.

### 4.3 Artificial Neural Network

Artificial Neural Networks have emerged as a popular algorithm for parameter estimation in PV cells. ANNs utilise interconnected nodes or neurons to process information and generate outputs. These networks are trained using a dataset of known input-output pairs to develop a mathematical model of the system. ANNs have been successfully implemented in estimating parameters of PV cells.

The implementation of ANNs in parameter identification of PV modules has shown promising results, with improved accuracy compared to conventional methods. ANNs' ability to learn complex non-linear relationships and adapt to changing input conditions make them a powerful tool for PV cell modelling and simulation.

Additionally, ANNs can learn from data, without the need for prior knowledge or assumptions about the PV cell. This makes them more adaptable to different types of PV cells and environmental conditions, leading to more accurate parameter estimation. ANNs can handle noisy and incomplete data, making them more robust to measurement errors and environmental variations.

#### 4.3.1 Proposed ANN Architecture Approach

For building the ANN model, a large dataset of input-output pairs is a prerequisite for modelling, where the input is the irradiance and temperature and the output is the measured MPP of the solar cell. This dataset is applied to train the ANN.

One of the primary tasks in building an ANN is to identify the best hyperparameters for the model. These hyperparameters include the architecture of the ANN, such as total number of input and hidden layers and each layer's neuron selection, the activation function used, and other important factors that affect the performance of the model.

Total number of input and hidden layers and each layer's neuron selection should be chosen based on the complexity of the problem and the size of the dataset. The activation function used in each layer should also be carefully chosen.

The learning rate, batch size, and number of epochs are the other hyperparameters required to define the ANN architecture. These parameters impact the efficiency of the model, therefore should be chosen to perfect the model of ANN.

After defining the architecture, train the model using an optimization algorithm. During training, the ANN adjusts its weights and biases to minimise the gap between the predicted and the actual dataset. Optimization algorithm Adams can be used.

Once the ANN is trained, a separate set of input-output pairs (called the validation set) is used to evaluate its performance. This is important to ensure that the ANN is not overfitting to the training data and can generalise well to new inputs.

Once the ANN is trained and achieves a high level of accuracy, the parameters of the model can be estimated using the weights and bias of each layer.

These parameters are then used to map new input data using a trial method based on ranges of parameters needed. It is important to note that this process should be done carefully and with proper validation to ensure the accuracy and reliability of the ANN's outputs.

Comparative Study of MPPT and Parameter Estimation of PV Cells using ANN

### 4.3.2 Overview of ANN Training

To train the ANN model, the appropriate value of hyperparameters needs to be identified for building the architecture. The training dataset is split into 80 percent of training and 20 percent testing. Keras from tensorflow library is used to define the neurons and number of layers associated with it. Input layer with shape = 2 is defined along with 2 hidden layers and activation function as ReLU throughout. Adam is an optimizer and MSE is considered for monitoring the loss during training.

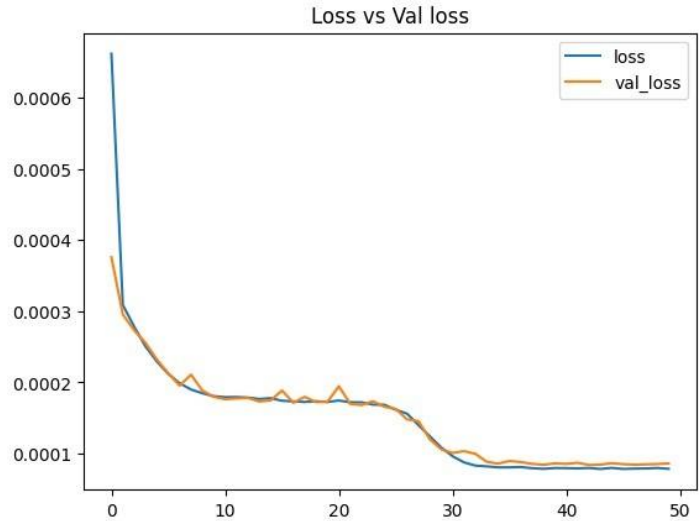

Figure 7: Fig 7. Loss during ANN training

In Figure 7, It is noticed that after 30 Epoch, loss is reduced to less than 0.0001. This indicates the successful implementation of the model.

### 4.3.3 Objective function

To measure the efficiency of model, such as ANN model, mean error, mean square error, and standard deviation is used as error calculation.
 where Ppredi is the ith power calculated by ANN, Pmeasi is the ith power by measured value, and u is the mean of the

$$ME = \frac{1}{N}\sum_{i=1}^{N}\left(P_{meas_i} - P_{pred_i}\right)$$

$$MSE = \frac{1}{N}\sum_{i=1}^{N}\left(P_{meas_i} - P_{pred_i}\right)^2$$

$$\sigma = \sqrt{\frac{1}{N}\sum_{i=1}^{N}\left(P_{math_i} - \mu\right)^2}$$

Figure 8: Objective function formula

measured data.

Comparative Study of MPPT and Parameter Estimation of PV Cells using ANN

# 5 Result

Upon implementing the ANN model, the error and gaps between the measured output and the modelled output decreased significantly.Refer Table 4 and 5.

The objective function applied to measure the error before and after the training gives the following outputs.
Table 3 Absolute Error and MPP values before and after training the Dataset

| Index | Irradiance | Temp | Pmpp | Pmpp | Absolute error | Pmpp | Absolute error |
|---|---|---|---|---|---|---|---|
| 1 | 758.20481 | 25 | 170.1704039 | 130.1439 | 40.02648 | 170.823415 | 0.6530116 |
| 2 | 1052.9782 | 25 | 226.5815128 | 173.1067 | 53.47478 | 225.512129 | 1.069383 |
| 3 | 690.8158 | 25 | 135.3307771 | 104.3459 | 30.98481 | 136.376119 | 1.0453419 |
| 4 | 354.3819 | 32.7525 | 79.01153151 | 60.16712 | 18.84440 | 78.6257803 | 0.3857511 |
| 5 | 5.8948 | 24.7592 | 1.12620655 | 0.8648653 | 0.261341 | 1.13762379 | 0.0114172 |
| 6 | 337.1830 | 32.2157 | 67.92064731 | 53.64569 | 14.27495 | 70.0431348 | 2.122487 |
| 7 | 312.8262 | 31.3557 | 68.09774599 | 51.83700 | 16.26073 | 67.7019406 | 0.3958053 |
| 8 | 337.8380 | 39.8892 | 93.5353097 | 74.41641 | 19.11889 | 97.1195477 | 3.584238 |
| 9 | 748.6899 | 25 | 140.9083832 | 108.8447 | 32.06360 | 142.312376 | 1.403993 |
| 10 | 270.4983 | 34.6003 | 53.8320397 | 33.73084 | 20.10119 | 44.0441940 | 9.78784 |
| 11 | 26.7269 | 24.3589 | 5.599556093 | 4.315890 | 1.283665 | 5.67068945 | 0.0711333 |
| 12 | 25.0901 | 18.0228 | 5.427678173 | 4.126344 | 1.301333 | 5.4214532 | 0.006224 |
| 13 | 28.5818 | 22.6388 | 6.065162425 | 4.61099 | 1.45417 | 6.05839301 | 0.0067694 |
| 14 | 509.8802 | 34.189 | 136.9420536 | 105.8897 | 31.0522 | 138.811745 | 1.86969 |
| 15 | 866.5382 | 25 | 209.5477248 | 159.9034 | 49.6442 | 208.8015 | 10.746213 |
| 16 | 167.7977 | 27.443 | 36.18891578 | 27.80469 | 8.38421 | 36.3949765 | 0.20606 |
| 17 | 688.2516 | 38.7823 | 158.0277967 | 120.9957 | 37.0320 | 158.713578 | 0.685782 |
| 18 | 336.2404 | 38.6242 | 68.70395999 | 52.65637 | 16.0475 | 68.7156330 | 0.0116730 |
| 19 | 40.3040 | 25 | 8.637984468 | 6.701487 | 1.93649 | 8.80011887 | 0.162134 |

Source: NALDEO Technologies Industries

Table 4. Errors before ANN training

| Parameter | Value |
|---|---|
| MSE | 638.92733514 |
| RMSE | 25.27701198 |
| R2 Score | 0.869586991 |
| Standard Deviation | 16.331738628 |
| Mean error | 19.29 |

Table 5. Errors after ANN training

| Parameter | Value |
|---|---|
| MSE | 4.4274335522 |
| RMSE | 2.104146751 |
| R2 Score | 0.9990984 |
| Standard Deviation | 1.95332 |
| Mean error | -0.78 |

Table 6. Parameter estimation before ANN training

Comparative Study of MPPT and Parameter Estimation of PV Cells using ANN

| Parameter | Value |
|---:|:---:|
| Lid | 0.15 |
| u-c | 20 |
| u-v | 0 |
| eta-m | 0.1 |
| iam-b | 0.05 |
| albedo | 0.2 |
| wiring | 0.02 |
| soiling | 0.02 |

Table 7. Parameter estimation after ANN training

| Parameter | Value |
|---:|:---:|
| Lid | 0.08605087 |
| u-c | 19 |
| u-v | 0 |
| eta-m | 0.6705674 |
| iam-b | 0.088797617 |
| albedo | 0.97092277 |
| wiring | -0.09991056 |
| soiling | -0.09991056 |

To evaluate the accuracy of a model, a graph can be plotted showing the measured and modelled output versus time, typically over a period of 10 days. Refer Figure 9 and 10. By comparing this graph before and after training the model, it is possible to determine whether the model's accuracy has improved.

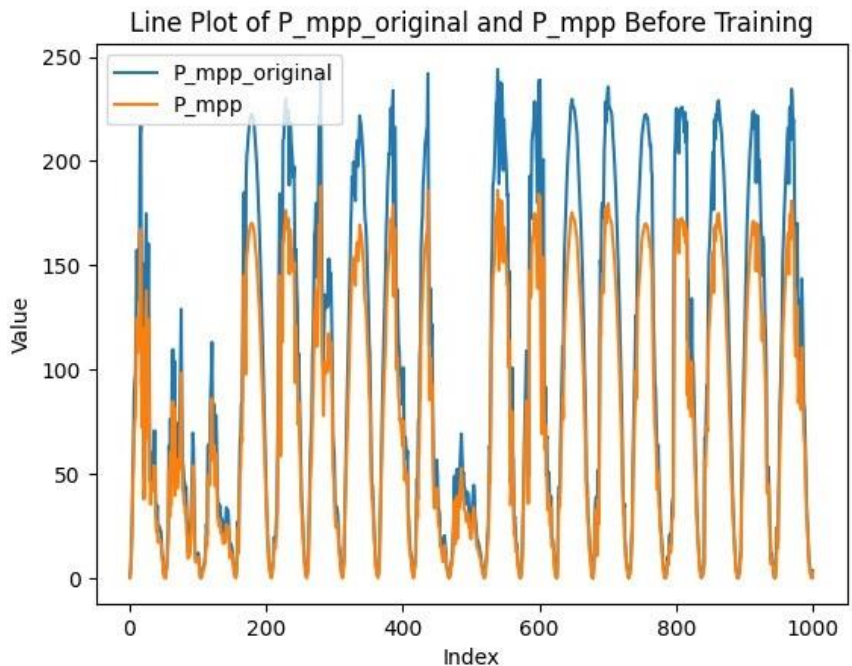

Figure 9: Line plot of P-mpp-original and P-mpp before training

If the accuracy has improved, this indicates that the model's parameters have been effectively tuned and the model can now accurately predict the peak power output even when presented with previously unseen input data. This is an important result, as it indicates that the model is robust and reliable, and can be trusted to provide accurate predictions in a variety of situations.

Comparative Study of MPPT and Parameter Estimation of PV Cells using ANN

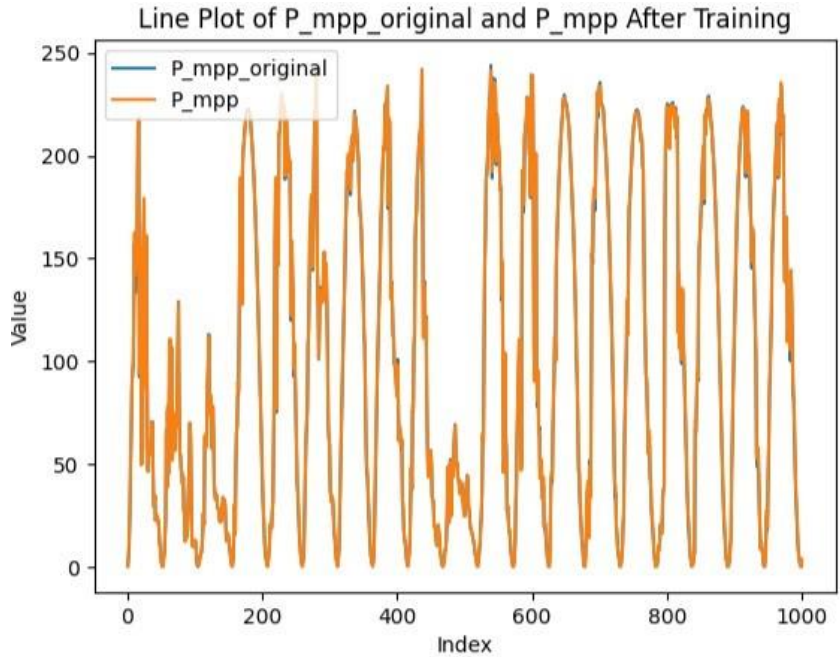

Figure 10: Line plot of P-mpp-original and P-mpp after training

## 5.1 Performance Evaluation

To evaluate the efficiency of the built model for its stability and accuracy, Performance evaluation technique is needed. For this, a Bland-Altman test is conducted. It's a statistical analysis test that compares the predicted and the measured values. The difference is then checked. If it is within u±2a i.e. 95 percent limits of acceptability, the model is said to be efficient model.

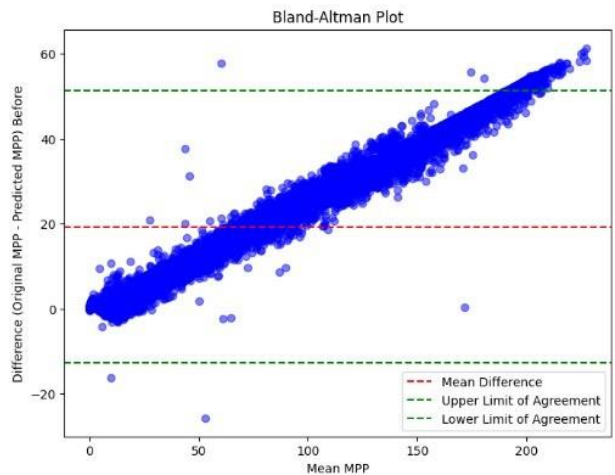

Figure 11: Bland-Altman Test

Most of the data points lie inside the limit of agreement and is more than 95 percent, this confirms the accuracy of the model

Bibliography.

# 6 Conclusion

In this paper, ANN ML based method for parameter estimation of PV system models is applied, considering the Irradiance and Temperature as its input pair and Power as output. The coefficient of determination output was measured to be higher than 99 percent in constant state to give the verdict of an efficient ANN Model. The ANN approach has produced better values for parameter estimation than meta heuristic algorithms and fuzzy logic Algorithms. In Addition,the output shows higher precision in mapping the parameters value, which in response, decreased the mean squared error significantly. In this paper, a brief equation of the PVlib model of the solar cell has also been presented.